\documentclass[sigconf]{acmart}

\AtBeginDocument{%
  \providecommand\BibTeX{{%
    \normalfont B\kern-0.5em{\scshape i\kern-0.25em b}\kern-0.8em\TeX}}}

\copyrightyear{2020}
\acmYear{2020}
\setcopyright{rightsretained}
\acmConference[KDD '20]{Proceedings of the 26th ACM SIGKDD
Conference on Knowledge Discovery and Data Mining}{August 23--27,
2020}{Virtual Event, CA, USA}
\acmBooktitle{Proceedings of the 26th ACM SIGKDD Conference on
Knowledge Discovery and Data Mining (KDD '20), August 23--27, 2020,
Virtual Event, CA, USA}
\acmDOI{10.1145/3394486.3403382}
\acmISBN{978-1-4503-7998-4/20/08}

\begin{document}
\fancyhead{}

\title{Bootstrapping Complete The Look at Pinterest}

\author{Eileen Li, Eric Kim, Andrew Zhai, Josh Beal, Kunlong Gu}
\affiliation{Visual Search, Pinterest}
\email{{eileenli,erickim,andrew,jbeal,kgu}@pinterest.com}


\begin{abstract}

Putting together an ideal outfit is a process that involves creativity and style intuition. This makes it a particularly difficult task to automate. Existing styling products generally involve human specialists and a highly curated set of fashion items. In this paper, we will describe how we bootstrapped the Complete The Look (CTL) system at Pinterest. This is a technology that aims to learn the subjective task of ``style compatibility'' in order to recommend complementary items that complete an outfit. In particular, we want to show recommendations from other categories that are compatible with an item of interest. For example, what are some heels that go well with this cocktail dress? We will introduce our outfit dataset of over 1 million outfits and 4 million objects, a subset of which we will make available to the research community, and describe the pipeline used to obtain and refresh this dataset. Furthermore, we will describe how we evaluate this subjective task and compare model performance across multiple training methods. Lastly, we will share our lessons going from experimentation to working prototype, and how to mitigate failure modes in the production environment. Our work represents one of the first examples of an industrial-scale solution for compatibility-based fashion recommendation.

\end{abstract}



\begin{teaserfigure}
\begin{center}
\includegraphics[width=0.9\linewidth]{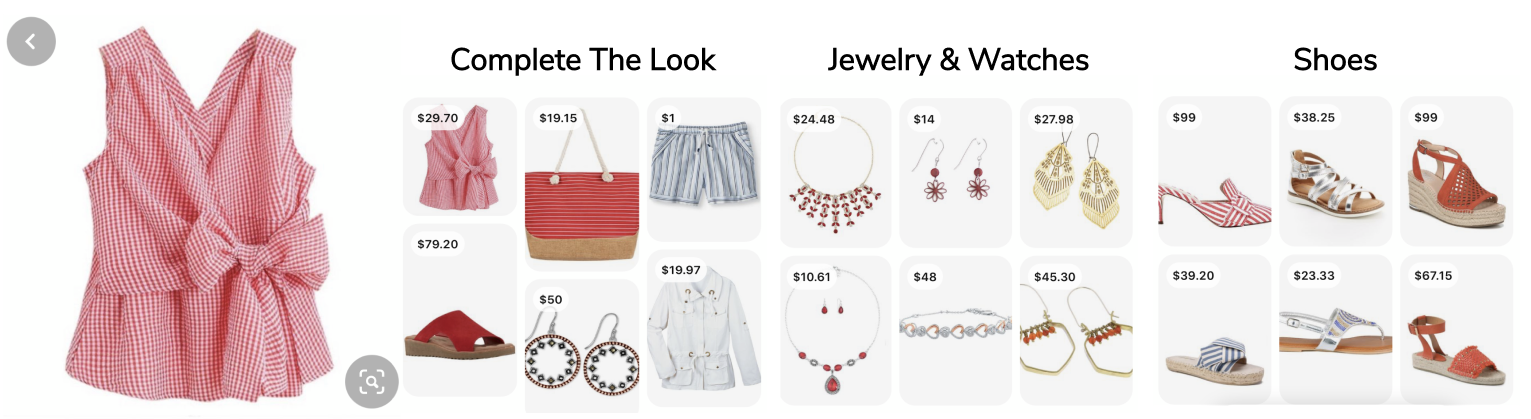}
\end{center}
  \caption{Complete The Look gives users outfit ideas and helps them find complementary products in related categories. The screen-captures above show outfit, jewelry, and shoes recommendations for the red striped blouse.}
\label{fig:application}
\end{teaserfigure}

\maketitle
\section{Introduction}

Since the number of online shoppers has grown exponentially in the past decade, platforms such as Amazon, Instagram, Taobao and Pinterest have all worked to create products that add value to the shopping experience.

In this crowded arena, Pinterest uniquely focuses on discovery and inspiration. Over 350M users visit the Pinterest website every month to discover new ideas in the realm of fashion, beauty, food and drinks, travel, home decor, and more. On Pinterest, they discover and save ``pins'', images with rich metadata attached, such as title, description, and url. Pinterest's ultimate goal is to turn inspiration into real-life actions and improvements. In recent years, visual search products such as Shop The Look~\cite{shop-the-look} helped to close this loop by enabling users to make purchases directly from pins they have saved. Shop The Look uses the Unified Embedding model~\cite{unified-embedding} trained on millions of pieces of Pinterest content to find products that are visually similar to the objects detected in the image. 

As a follow-on to Shop The Look, we have been working on a novel shopping experience called Complete The Look (CTL) (see Figure~\ref{fig:application}). Whereas Shop The Look has the goal of visual exact match and retrieval, Complete The Look aims to find products that are visual complements. CTL helps users find ideas about how to style a particular product and gives them the option to continue their shopping experience in related categories. We hope to answer queries such as, ``What are some hats that I might buy to wear with this jacket?''.

Completing an outfit is not a new problem, but existing solutions require fashion products that have been highly curated and matched by human stylists. There are also systems that use past engagement to build complementary recommendations (i.e., ``You liked this item? Maybe you'll like this other item too.''). In contrast, we describe a solution for when such explicit user engagement does not exist. Specifically, we offer our approach to \textbf{bootstrapping the Complete-the-Look system} and how we made it work with our diverse corpus of tens of millions of products. To do so requires answering some challenging questions, such as:

\begin{itemize}
  \item How do we generate an outfit dataset that is high-quality and relevant to the shopping content on Pinterest?
  \item How do we evaluate this subjective task of ``style compatibility''?
  \item How do we handle mislabeled or missing product metadata?
  \item How do we design a system that is performant, scalable, and easy to maintain?
\end{itemize}
In particular, our contributions are:

\begin{enumerate}
    \item We describe an automatic way of generating an outfit dataset from the Pinterest platform of over 1M unique outfits and 4M fashion items, leveraging existing technologies in object detection, image style classification and attribute classification. To the best of our knowledge, this is the largest known dataset of fashion outfits, and we will release 100K outfits from our training set with our entire test set of 25K outfits.
    \item We combine this dataset with a Convolutional Neural Net (CNN) to learn useful style embeddings. We present quantitative experimental results comparing performance across multiple training methodologies, including loss functions (contrastive, triplet, classification), data preprocessing, network architecture, and dataset collection.
    \item We deploy and evaluate this model in an end-to-end recommendation system that performs retrieval from a diverse corpus of Pinterest shopping products, overcoming challenges in serving infrastructure, domain adaption, and metadata mislabeling.
\end{enumerate}

\section{Related Works}

\subsection{Visual Similarity}
Fine-grained visual similarity and retrieval is a well-studied problem. The earlier approaches for modeling visual similarity relied mostly on hand-crafted features and attributes~\cite{whittlesearch}~\cite{color-matching}. More recent approaches using CNNs have been able to achieve state-of-the-art results~\cite{visual-sim}~\cite{deep-fashion}~\cite{WhereToBuyItICCV15}~\cite{cross-domain}. The challenge falls on curating a high-quality matching dataset (i.e., scene with bounding box and object pairs). The model trained on this dataset, usually by way of metric learning, learns to transform an image to its embedding representation. During retrieval, a query embedding is compared with many candidate embeddings to find the most similar results. This technology powers much of Pinterest's visual search~\cite{DBLP:journals/corr/JingLKZXDT15}, and is the backbone to the system on which CTL is built.

\subsection{Style Modeling}
Earlier approaches in style modeling relied on crowdsourcing hand-labeled datasets~\cite{hipster-wars}. This involved having explicit buckets for each style (e.g., ``Bohemian'' vs. ``Classy''), which is limiting due to its subjective nature. Since then, there have been efforts in scaling the annotations for fashion datasets~\cite{weak-data}~\cite{street-style} in order to enable much more fine-grained attribute classification. \cite{latent-style} uses the topic modeling approach to discover latent ``styles'', but still requires a hand-crafted set of attributes. More recently, researchers have used outfits from the popular site Polyvore to train models such as SiameseCNN~\cite{siamese}, sequence models (e.g., LSTM~\cite{lstm}), and graph-inspired networks~\cite{graph} for learning style~\cite{style2vec}~\cite{capsule}~\cite{type-aware}~\cite{across-categories}. In contrast to our system, most of these approaches are not optimized for retrieval but are instead trained for fashion compatibility classification.

\subsection{Recommender Systems}
Recommendation systems for fashion have grown in importance as as increasing percentage of shopping moves online. The earlier systems used traditional machine learning methods (e.g., SVM, logistic regression) trained on a hand-annotated dataset and retrieved recommendations from a limited corpus~\cite{magic-closet}~\cite{image-rec}. More recently, Pinterest published a paper that focused on scene-to-product recommendation \cite{ctl-scene2product}---e.g., given a beach scene, recommend products that are complementary. Here we present a modified approach that handles product-to-product recommendation. Many recommendation systems, such as Alibaba's iFashion \cite{pog}, use past user engagement to train complementary models \cite{fashion-mining}~\cite{siamese}. We attempt the challenging task of bootstrapping such a system without explicit user data.
\section{System overview}

\begin{figure}[ht!]
\centering
\includegraphics[width=0.9\linewidth]{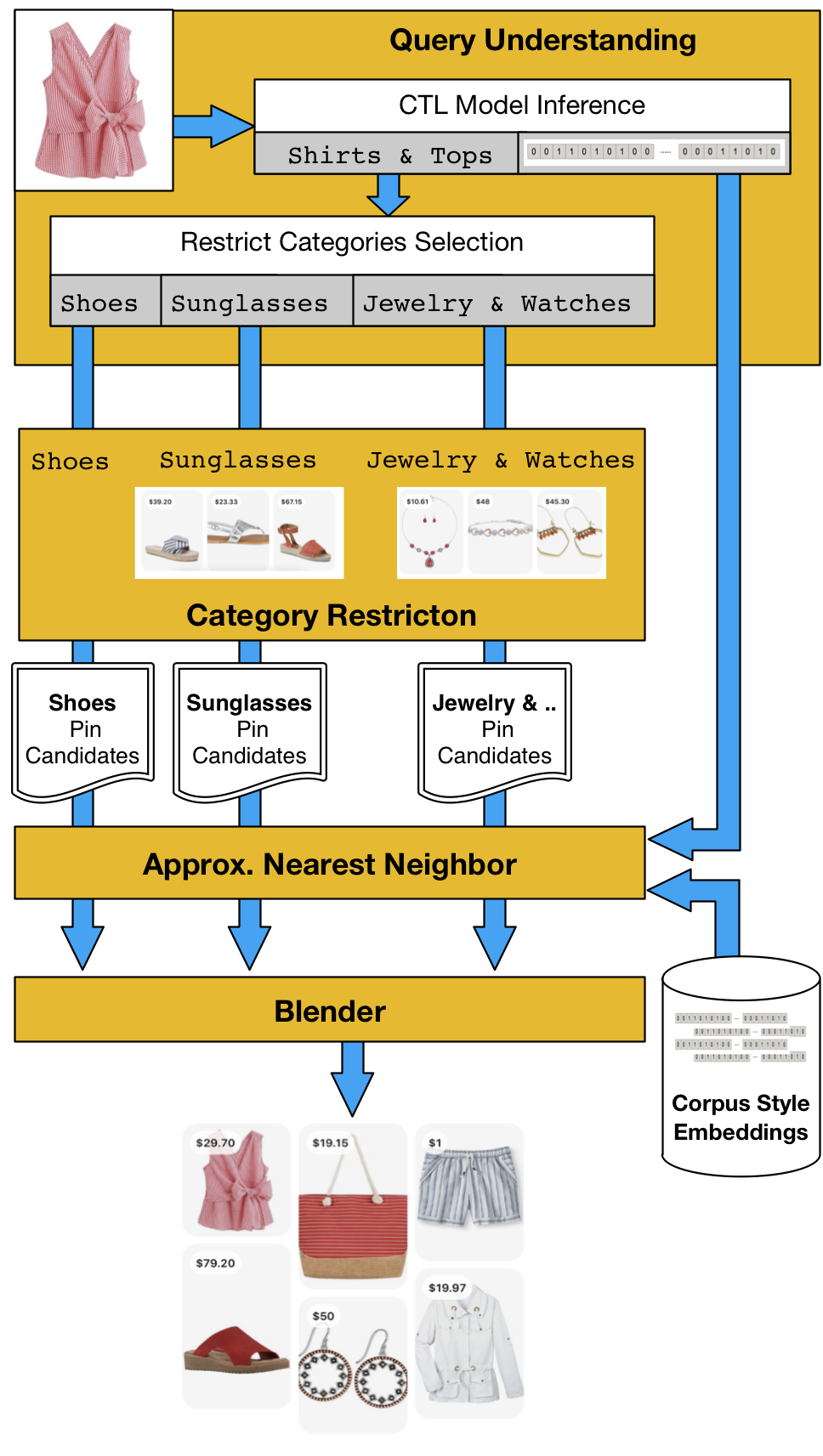}
\caption{Serving system overview for Complete The Look}
\label{fig:system_overview}
\vspace{-2em}
\end{figure}

We define Complete The Look (CTL) as the problem of matching a single product to multiple complementary products. While a user is viewing an apparel product on Pinterest, we aim to offer them complementary products to help them ``complete'' their look. The serving system is real-time, generating recommendations in a fraction of a second through four stages: \emph{query understanding}, \emph{candidate generation}, \emph{full scoring}, and \emph{blending} as shown in Figure~\ref{fig:system_overview}.

\textbf{Query understanding} takes the apparel product a user is currently viewing and enriches the query with additional features. For CTL, the core features are inferred from the CTL model, discussed more in Section~\ref{sec:model_arch}, which predicts the product category of the query along with the style embedding. The product category prediction is then expanded into the outfit apparel categories that are complementary. As an example, for the ``Shirts \& Tops'' query category, complementary categories to ``complete'' the look may be ``Shoes'', ``Sunglasses'', and ``Jewelry \& Watches''.

\textbf{Candidate generation} leverages the complementary categories from \emph{query understanding} to restrict our apparel corpus of millions of products to the items that match the given categories. Each category has its own list of candidates. Note that the corpus categories are also generated from offline batch inference of the CTL model and the inverted indices per corpus categories are built offline.

\textbf{Full scoring} takes each complementary category's candidate lists and runs approximate nearest neighbor (ANN) search of the query's style embedding against the complementary item's style embeddings to rank the results within a category. These ANN indices are built offline for each inverted index and contain the CTL model predicted style embeddings for the candidates.

\textbf{Blending} merges each complementary category's full-scored candidate lists and selects the final ordering to present to users.

\section{Dataset}
Fashion datasets such as Fashion136K~\cite{fashion136k} and StreetStyle~\cite{street-style} generally included ``in-the-wild'' street photos of people wearing clothes. More recently, composition outfit datasets from Polyvore website have been used for learning fashion tasks, such as FashionVC (20,726 outfits)~\cite{neurostylist}, Maryland Polyvore (33,375 outfits)~\cite{lstm}, and Polyvore Outfits (68,306 outfits)~\cite{type-aware}. Polyvore was a popular website where people created and shared collages of outfits. Each outfit image consisted of multiple fashion items that came together to form a cohesive style. Unfortunately, the site was shut down in 2018.


Pinterest is a visual discovery engine with billions of pieces of diverse content saved by people around the world. In order to build a product that works well in this varied ecosystem, we need a reliable way to gather a large-scale outfit dataset for training that can be refreshed periodically. By using an Image Style classifier and object detector, we have implemented an extraction flow to decompose outfit images into their set of fashion objects. We chose to filter for images in collage-like sets similar to those that were popular on Polyvore. These ``polyvore''-style images include objects that better match the domain of our product corpus, which typically consists of images with a clean background.

\begin{figure}[hbt!]
\includegraphics[width=1.0\linewidth]{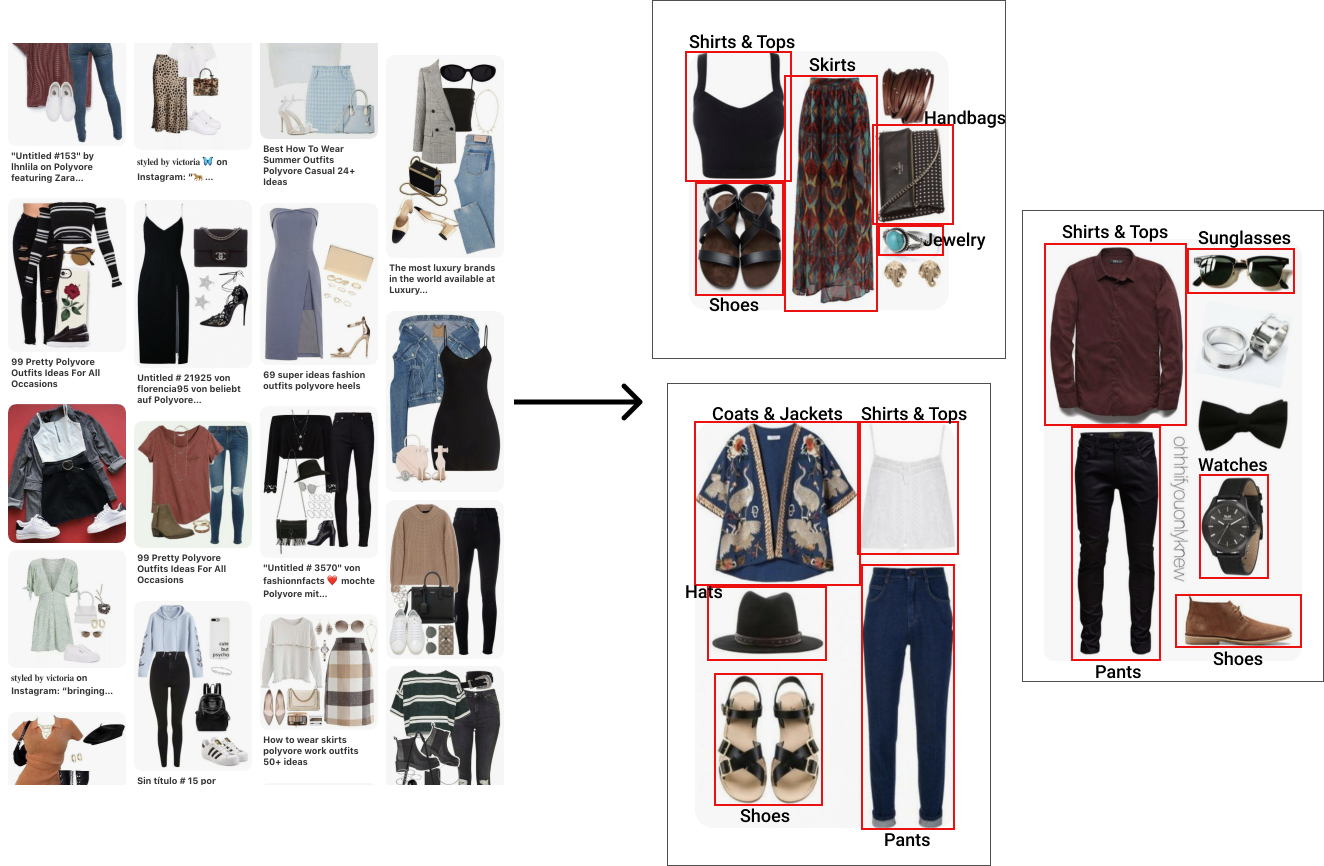}
\caption{Pinterest images are funneled through steps 1) fashion and ``polyvore''-style classification; 2) object detection and labeling; and 3) heuristics post-processing, resulting in high quality, high volume outfit dataset.}
\label{fig:dataset}
\vspace{-1.5em}
\end{figure}

At a high level, the dataset extraction flow is comprised of the following steps:
\begin{enumerate}
  \item We trained an Image Style classifier to identify ``polyvore''-style images on Pinterest. These images are not necessarily from polyvore.com, but simply follow the same visual style of an outfit collage (Figure~\ref{fig:dataset}).
  \item We ran our object detection model to gather bounding boxes and category labels for items on these ``polyvore''-style images.
  \item We cleaned up this dataset using a number of post-processing criteria for higher quality images. See Tables \ref{table:category_count} and \ref{table:items_count} for the dataset distributions broken down by category and number of items per outfit, respectively.
\end{enumerate}

\subsection{Image Style classifier}

\begin{figure}[ht!]
\includegraphics[width=1.0\linewidth]{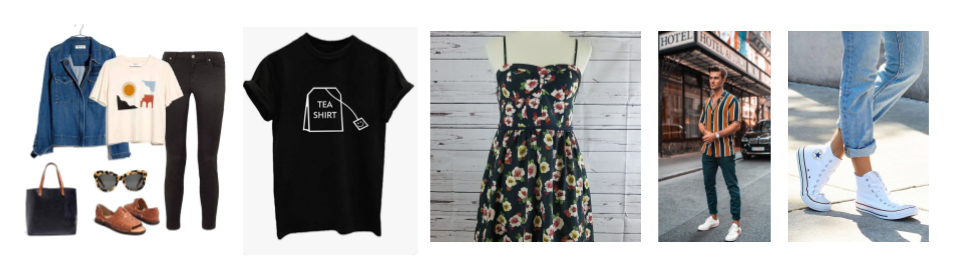}
\caption{Examples of Image Style categories for CTL model, from left to right: (1) Polyvore, (2) Product Shot, (3) Stock Photo, (4) Full Outfit, and (5) Cropped Outfit.}
\label{fig:shopping_style_categories}
\vspace{-1em}
\end{figure}

Image Style is a signal consisting of a family of multi-class and multi-label visual classifiers. Each classifier uses the Unified Embedding \cite{unified-embedding} as input and outputs a score between 0 and 1 for every style class. These classifiers are trained on human-curated image datasets with low label noise. There are five labels of interest for the CTL model: Polyvore, Product Shot, Stock Photo, Full Outfit, and Cropped Outfit (see Figure~\ref{fig:shopping_style_categories}). These labels are used to distinguish common types of shopping content in the fashion domain. By filtering on the Polyvore score at a high-precision threshold of 0.9, around 35M images in the fashion domain are obtained.

\subsubsection{Domain difference}
\begin{figure}[h!]
\includegraphics[width=0.6\linewidth]{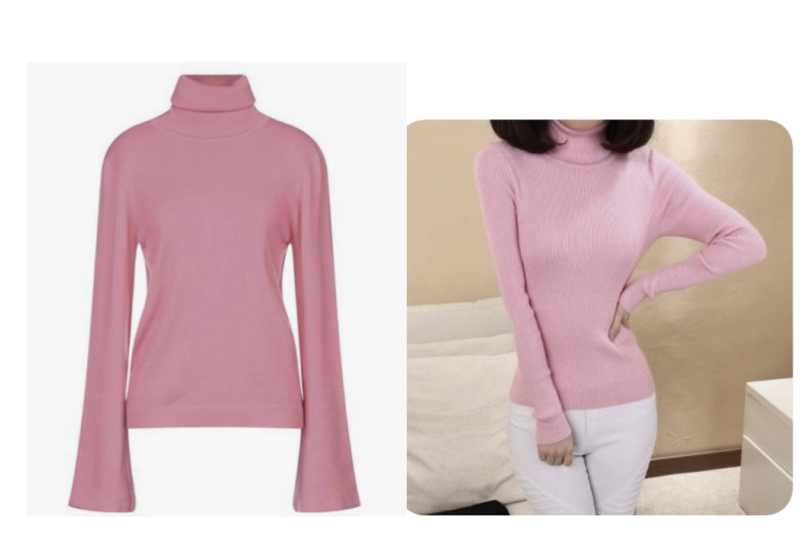}
\caption{(Left) Product Shot pins that match the image domain of our training set. We restrict CTL to only these images. (Right) Cropped Outfit pins that we filter out for CTL.}
\label{fig:single_product_vs_in_situ_image}
\vspace{-1em}
\end{figure}

We use the ``polyvore''-style to collect training data because objects detected from these images resemble products on solid backgrounds (i.e., Product Shot). As expected, quality decreases dramatically for out-of-domain products (i.e., Stock Photo, Full Outfit, Cropped Outfit). To ensure quality, we restrict CTL queries and results using the Image Style classifier to filter for Product Shots only (see Figure~\ref{fig:single_product_vs_in_situ_image}).

\subsection{Object Detection}
To decompose each ``polyvore''-style image into its individual articles of clothing, e.g., ``Shirts \& Tops'' and ``Pants'' (Figure~\ref{fig:dataset}), we utilize an object detection model that is trained on fashion categories.
Specifically, we use a Faster-RCNN \cite{DBLP:journals/corr/RenHG015} with a ResNeXt101 \cite{DBLP:journals/corr/XieGDTH16} backbone and Feature Pyramid Networks \cite{DBLP:journals/corr/LinDGHHB16}, trained using the Detectron \cite{Detectron2018} framework.

Our detection training set consists of 251,000 training images with 727,000 bounding boxes in the home decor and fashion domains.
We set aside a validation set consisting of 10,000 images with 30,000 bounding boxes.

At inference time, we use a whitelist of 21 fashion categories (Table~\ref{table:category_count}) to restrict the bounding boxes to the CTL categories. Of the 35M "polyvore"-style images, about 10M of them have at least one CTL object detected.

\begin{table}[h!]
\begin{center}
\begin{tabular}{||r|r|r||}
\hline
mAP & Precision & Recall \\ [0.5ex] 
\hline
77.2 & 76.0 & 82.4 \\
\hline
\end{tabular}
\end{center}
\caption{Detection evaluation metrics on our in-house test set of 10k fashion images. ``mAP'' is mean average precision. Precision and recall are obtained by choosing the operating point that maximizes the F1 score.}
\label{tab:dtn_eval_offline}
\vspace{-2em}
\end{table}

To assess the quality of the detection model, we collected an in-house test set of 10k images in the fashion category (see Table~\ref{tab:dtn_eval_offline}).
We found that the performance of the object detection model was satisfactory for the CTL dataset generation pipeline.

\begin{table}[h!]
\centering
\begin{tabular}{||l|r|r|r||} 
 \hline
 Label & \#Items(full) & \#Items(100K) & \#Items(test) \\ [0.5ex] 
 \hline
 Shoes & 856,775 & 94,059 & 22,728 \\ 
 Handbags & 810,501 & 88,955 & 20,729 \\
 Shirts \& Tops & 519,625 & 57,250 & 14,451 \\
 Pants & 468,845 & 51,706 & 11,666 \\
 Coats \& Jackets & 397,353 & 43,552 & 9,698 \\
 Dresses & 254,806 & 28,066 & 6,903 \\
 Jewelry & 222,743 & 24,409 & 7,813 \\
 Hats & 168,394 & 18,581 & 4,343 \\
 Skirts & 138,343 & 15,152 & 3,630 \\
 Sunglasses & 86,802 & 9,343 & 2,174 \\
 Shorts & 84,743 & 9,265 & 2,166 \\
 Scarves \& Shawls & 52,233 & 5,751 & 1,218 \\
 Watches & 51,613 & 5,605 & 1,313 \\
 \hline
\end{tabular}
\caption{Number of items in full, released, and test datasets from top 13 out of 21 total categories.}
\label{table:category_count}
\vspace{-3.5em}
\end{table}

\begin{table}[h!]
\centering
\begin{tabular}{||l|r|r|r||} 
 \hline
 \#Items & \#Outfits(full) & \#Outfits(100K) & \#Outfits(test) \\ [0.5ex] 
 \hline
 3 & 223,240 & 24,545 & 5,383 \\ 
 4 & 421,815 & 46,545 & 9,288 \\
 5 & 251,835 & 27,643 & 6,616 \\
 6 & 72,916 & 7,927 & 2,786 \\
 7 & 10,767 & 1,127 & 722 \\
 8 & 938 & 102 & 165 \\
 \hline
\end{tabular}
\caption{Number of clothing articles in an outfit; in full, 100K and test datasets.}
\label{table:items_count}
\vspace{-2em}
\end{table}

\subsection{Dataset Cleanup}
\label{sec:dataset_cleanup}

\begin{figure}[h!]
\includegraphics[width=0.8\linewidth]{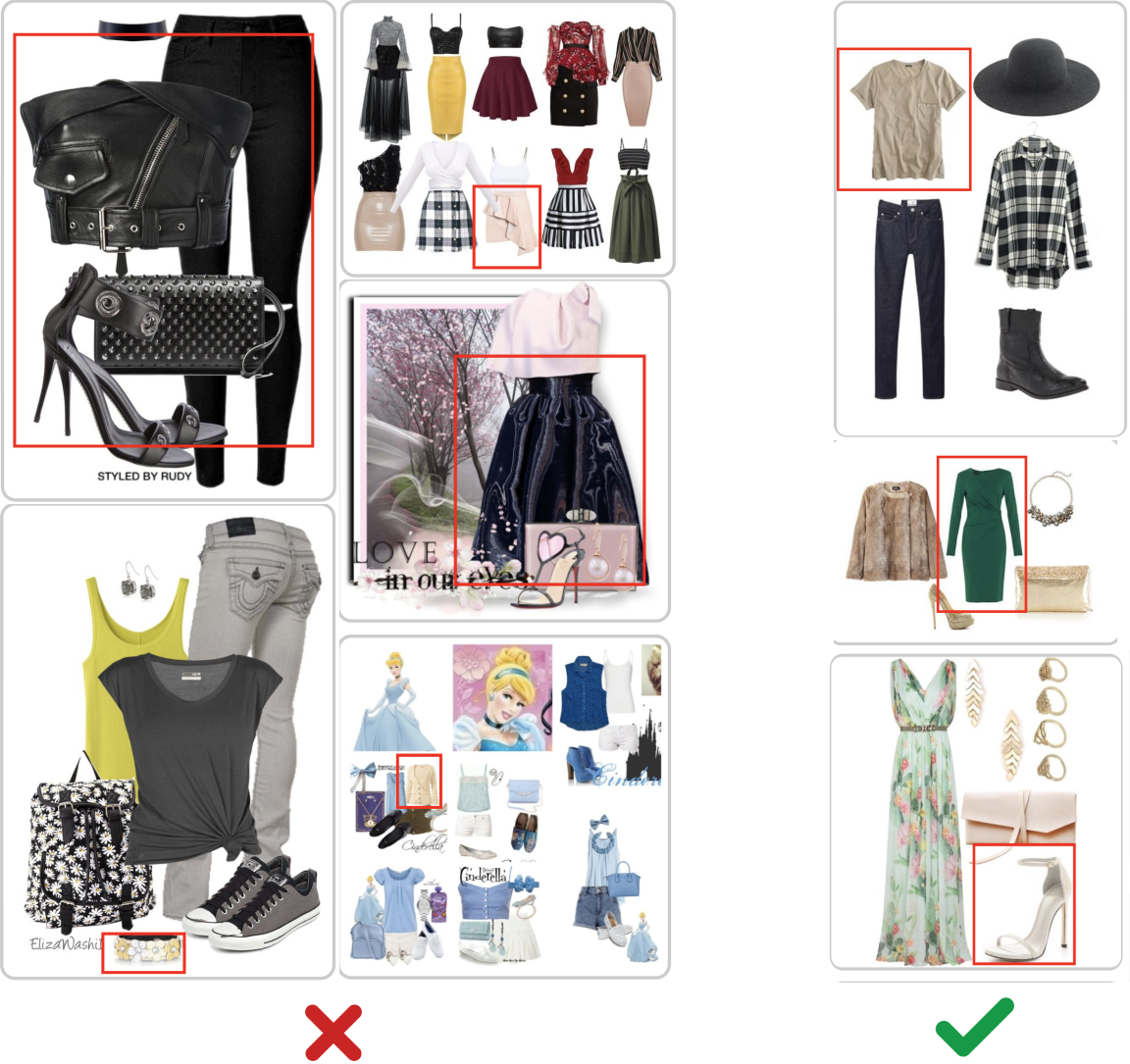}
\caption{We clean up the dataset using a number of heuristics to ensure clean bounding boxes with a cohesive set of fashion items.}
\label{fig:dataset_cleanup}
\vspace{-1em}
\end{figure}

We want outfits that have good style cohesion and high diversity in category and color. For example, training on a corpus consisting mostly of outfits with white tops and blue jeans may lead to a model that heavily biases towards recommending jeans for every query.
We also want crops with clean backgrounds so that they can better match the conditions of our product images. To help curate a higher quality dataset, we applied a series of post-processing steps:
\begin{itemize}
    \item Discard overlapping bounding boxes by applying non-maximum suppression \footnote{We use an IOU threshold of 0.1}.
    \item Discard bounding boxes that are too small, specifically less than 5\% of total image area.
    \item Enforce that each image has between 3 and 8 clothing items. We found that outfits outside of this range tend to be noisy and lack cohesion, e.g., consisting of several different outfits in the same image.
    \item Enforce at least 3 different types of clothing articles per outfit. This further guarantees that the image contains a complete outfit.
    \item Discard outfits where all items are of the same color. Many outfit images are monochromatic, resulting in models biased for matching color instead of style. Spot-checking results seem to show improvements with this filter, though the problem still exists (see qualitative feedback in Section \ref{human-eval}).
\end{itemize}

We tried curating by user engagement (e.g., clicks and repins) and image age (i.e., number of days on Pinterest) to get popular, trending outfits, but we did not find a noticeable difference in the generated outfits.

The final dataset has \textbf{1,006,519 outfits and 4,246,430 fashion objects}. To the best of our knowledge, this is the largest known dataset of fashion outfits. We keep 24,960 outfits (109,471 items) as a holdout set and train on the remaining 981,559 outfits (4,136,959 items). The female:male ratio is about 10:1. 100K outfits from the training set and the entire test set are made publicly available \footnote{\href{https://github.com/eileenforwhat/complete-the-look-dataset}{https://github.com/eileenforwhat/complete-the-look-dataset}}.
\section{Methodology}

\begin{figure}[ht!]
\centering
\includegraphics[width=1.0\linewidth]{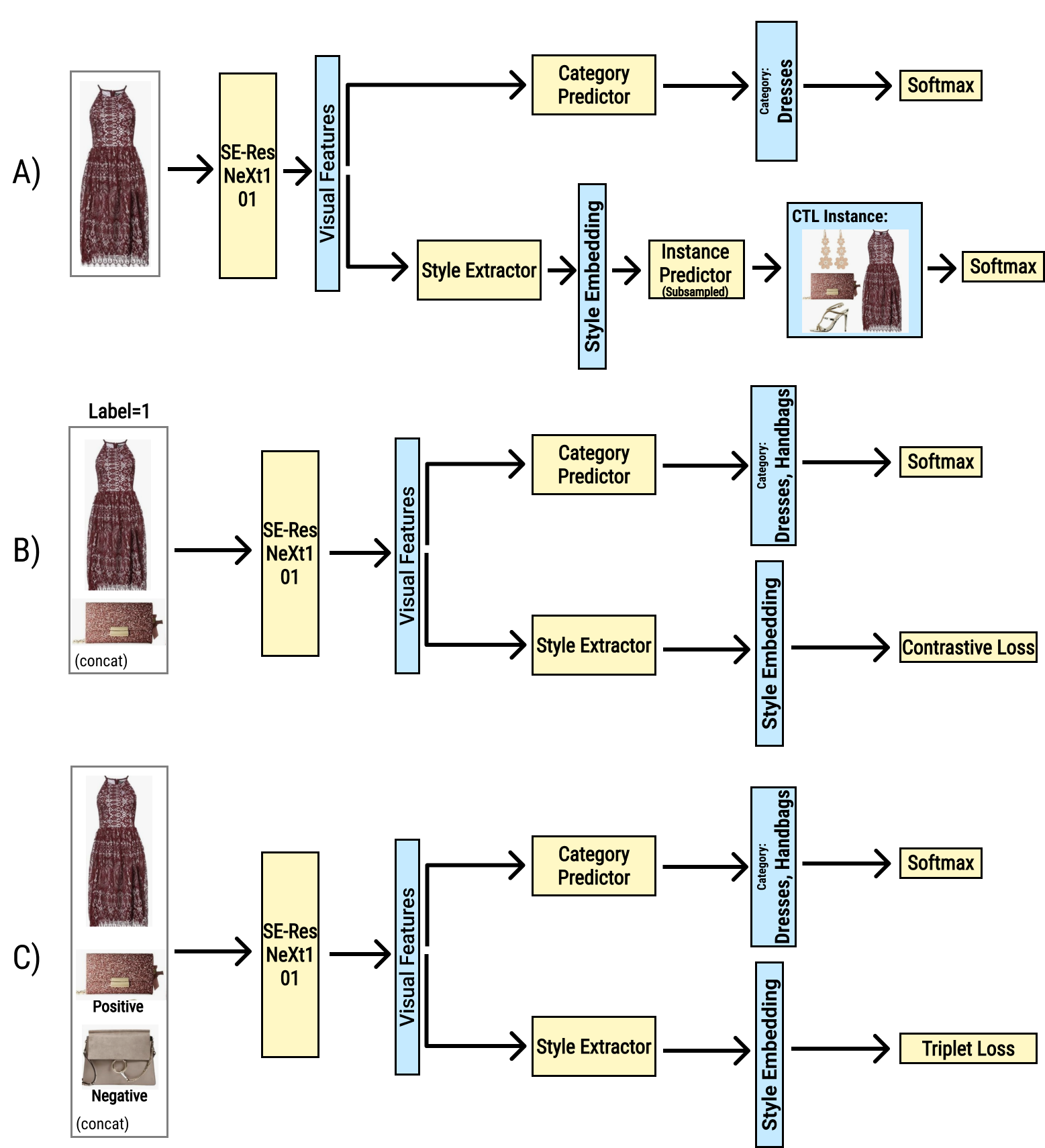}
\caption{The CTL model has three components: visual featurizer, category predictor, and style . We compare three different ways to train: A) classification ``proxy'' loss, B) contrastive loss, and C) triplet loss.}
\label{fig:model_arch}
\vspace{-1em}
\end{figure}

\subsection{Model Architecture}~\label{sec:model_arch}
The CTL model is comprised of three modules: the \emph{visual featurizer}, \emph{category classifier}, and \emph{style extractor}. We try to leverage as much existing Pinterest technology as possible, lowering the risk and maintenance cost of the system.

\textbf{The visual featurizer} uses the same SE-ResNext101 backbone \cite{DBLP:journals/corr/abs-1709-01507} as Pinterest's Unified Embedding \cite{unified-embedding}. Unified Embedding is a multi-task learning model that has been trained for three visual discovery tasks: Flashlight \cite{flashlight}, Lens \cite{lens}, and Shop The Look \cite{shop-the-look}. The Unified Embedding is the primary image-based embedding used by many systems at Pinterest.

``Layer4'' of the featurizer backbone is fed into two separate networks: one for category classification, and another for style learning.

\textbf{The category classifier} predicts the top category label for each input image, from a total of 21 fashion CTL categories. These predictions are important because we need to determine the query category (e.g., ``Dresses'') and also filter CTL results for particular candidate categories (e.g., ``Shoes'' and ``Coats \& Jackets''). At serving time, products are stored in our retrieval index keyed by the category prediction from this classifier, so we can efficiently retrieve and filter results on any set of product categories.

The ground truth labels for this classifier come from the detection model (see Table~\ref{tab:dtn_eval_offline} for detection evaluation metrics). The category classifier is implemented as two fully connected (FC) layers, achieving accuracy of 98.15\% across all categories on the test set.

\textbf{The style extractor}~\label{sec:style_learning} is a neural network with two FC-layers that outputs a 128-dimensional style embedding. We use batch normalization and a dropout rate of 0.5. We experimented with three different training methods: a classification network with ``proxy'' loss, contrastive loss (siamese) \cite{siamese} and triplet loss~\cite{triplet}. Our production CTL system uses the best performing model with triplet loss. See section~\ref{evaluation} for experiment results. 

The motivation for the classification network is its similarity to the Unified Embedding model \cite{unified-embedding}, if for example, we decide to add CTL as an additional task to our Unified Embedding training. In the classification model, each outfit has a unique instance label. At training time, we feed into the model the image of a fashion item. From the output layer of the style extractor, we calculate the softmax cross entropy loss for predicting into a subset of ``proxy'' instances. We sample the number of proxies to be 2048 which is consistent with Unified Embedding training. 

\begin{figure}[ht!]
\centering
\includegraphics[width=1.0\linewidth]{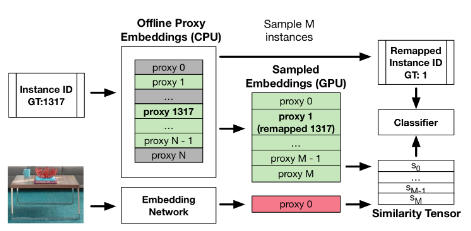}
\caption{We take this figure directly from the Unified Embedding paper~\cite{unified-embedding}. It depicts how we get a subset of ``proxy'' instances for calculating classification loss.}
\label{fig:proxy}
\vspace{-1em}
\end{figure}

To sample pairs for training a siamese network, we follow the sampling methodology in~\cite{siamese}. Similarly, we guarantee that each of our positive pairs comes from different categories (``heterogeneous dyads'') while negatives are randomly sampled regardless of category. This is done to dissuade learning visual similarity in place of style compatibility. We then optimize using contrastive loss:
\[\mathcal{L}_{contrast}(i, j) := y_{ij}D^2_{ij} + (1-y_{ij})[\alpha - D_{ij}]^2_+\]

where $D_{ij}$ denotes the distance between samples $i$ and $j$, and $y_{ij}$ is $1$ if samples $i$ and $j$ have the same label and $0$ otherwise.

We also train a triplets network by sampling \emph{(anchor\_image, pos\_image, neg\_image)}. This is similar to the scene-based CTL paper~\cite{ctl-scene2product}, where they trained a triplet network using \emph{(anchor\_scene, pos\_product, neg\_product)}. In our case, however, every item in the triplet is a crop of a product. We optimize using the loss function:
\[\mathcal{L}_{triplet}(a, p, n) := [D^2_{ap} - D^2_{an} + \alpha]_+\]

where $a$, $p$, $n$ denote the query, positive, and negative samples respectively; $D^2_{ap}$ denotes the distance between the anchor and positive samples, $D^2_{an}$ denotes the distance between the anchor and negative samples; and $\alpha$ is the margin term.

\subsection{Training and deployment details}
We train our models using the PyTorch framework~\cite{paszke2019pytorch} with 8 GPUs and the model takes about 10 hours to converge. We use Apex~\cite{apexamp} mixed precision to increase training efficiency. We use the Adam optimizer with learning rate of 0.048. For deployment, we deploy the PyTorch model to C++ directly by serializing to TorchScript.

\section{Evaluation} \label{evaluation}
\subsection{Methods}

Evaluating style compatibility is challenging because of its subjective nature. We compare our model performance using multiple methods to get a comprehensive assessment. We have a test set of 25K outfits (109,847 items), which we use to measure retrieval recall and Fill-in-the-Blank (FITB) accuracy. We also perform end-to-end evaluation on Pinterest's real product corpus, leveraging our in-house fashion specialists for labeling. Lastly, CTL is launched internally, allowing us to conduct user studies and gain valuable insights about how real users think when they use the product.

\subsubsection{Recall@\{1, 5, 10\}}
\label{sec:eval_recall_at_k}
We measure the retrieval recall on a sub-sampled corpus. Since CTL is ultimately used in a retrieval setting, measuring Recall@K (R@K) most closely mirrors the production task. 

In order to calculate exact recall, we limit the number of items per outfit in the test set to be exactly five, removing outfits that do not meet this criteria. For each item of an outfit, we calculate the R@\{1, 5, 10\} by retrieving the top matches using k-nearest neighbors, with euclidean distance between style embeddings as the distance measure. The retrieval corpus includes the other items from the same outfit (positives) and randomly sampled items (negatives). We calculate R@K according to two ways of generating the corpus: 1) sampled across all categories, and 2) restricted to one category at a time (and taking the mean). In both cases, the total size of the retrieval corpus is N=200. 
Thus, R@K is computed by the following:

    \[R@K= \dfrac{{(\#\;positives\;in\;top\;K)}}{min[(total\;num\;items\;in\;outfit) - 1, K]}\]
    \vspace{0.25em}

\begin{figure}[h!]
\centering
\includegraphics[width=1.0\linewidth]{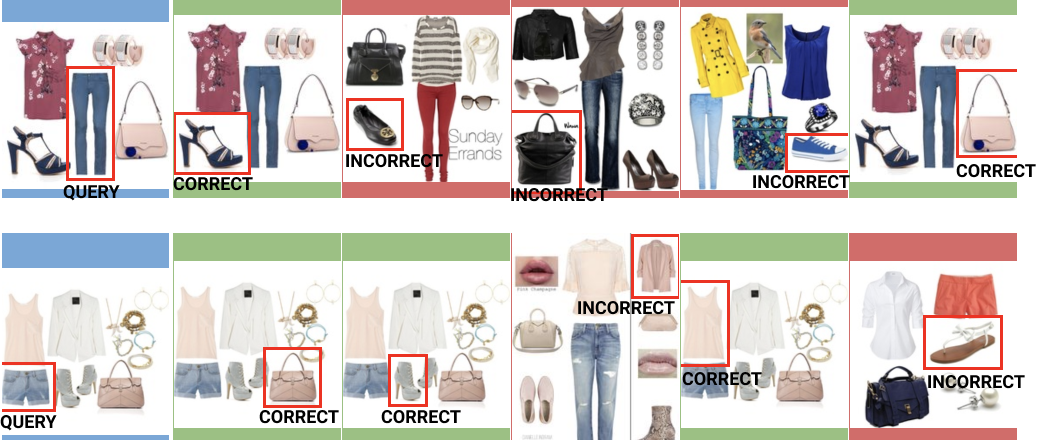}
\caption{Example of the retrieval evaluation. For each item in an outfit, we calculate R@\{1,5,10\}. The result is considered correct if it belongs to the same outfit as the query.}
\label{fig:retrieval}
\vspace{-1em}
\end{figure}

Since metrics gathered from both methods of sampling are highly correlated, we only report on the latter in this paper for simplicity. In Figure \ref{fig:retrieval}, we visualize some results from the retrieval task.


\subsubsection{Fill-in-the-Blank (FITB)}

For the FITB task, we remove one item from each outfit. The model then has to pick from the positive item and three randomly sampled items that belong to the same category. Unlike the retrieval task, FITB uses multiple products as queries rather than just a single product. Current SOTA methods for this task  (e.g., \cite{lstm}, \cite{type-aware}) are not suited for our large-scale one product to multi-complementary product retrieval.

A downside of both R@K and FITB evaluation tasks is that they only consider clothing articles from the same ``polyvore''-style image to be positive examples for a given query clothing article.
In fact, it's likely that clothing articles from other ``polyvore''-style images in the test corpus could be compatible with the query; for instance, blue jeans tend to be compatible with many other items. 
We address this shortcoming by using human evaluation (see Section~\ref{sec:eval_human_judgement}).

\subsubsection{Human Judgment}
\label{sec:eval_human_judgement}

\begin{figure}[htb!]
\centering
\includegraphics[width=0.8\linewidth]{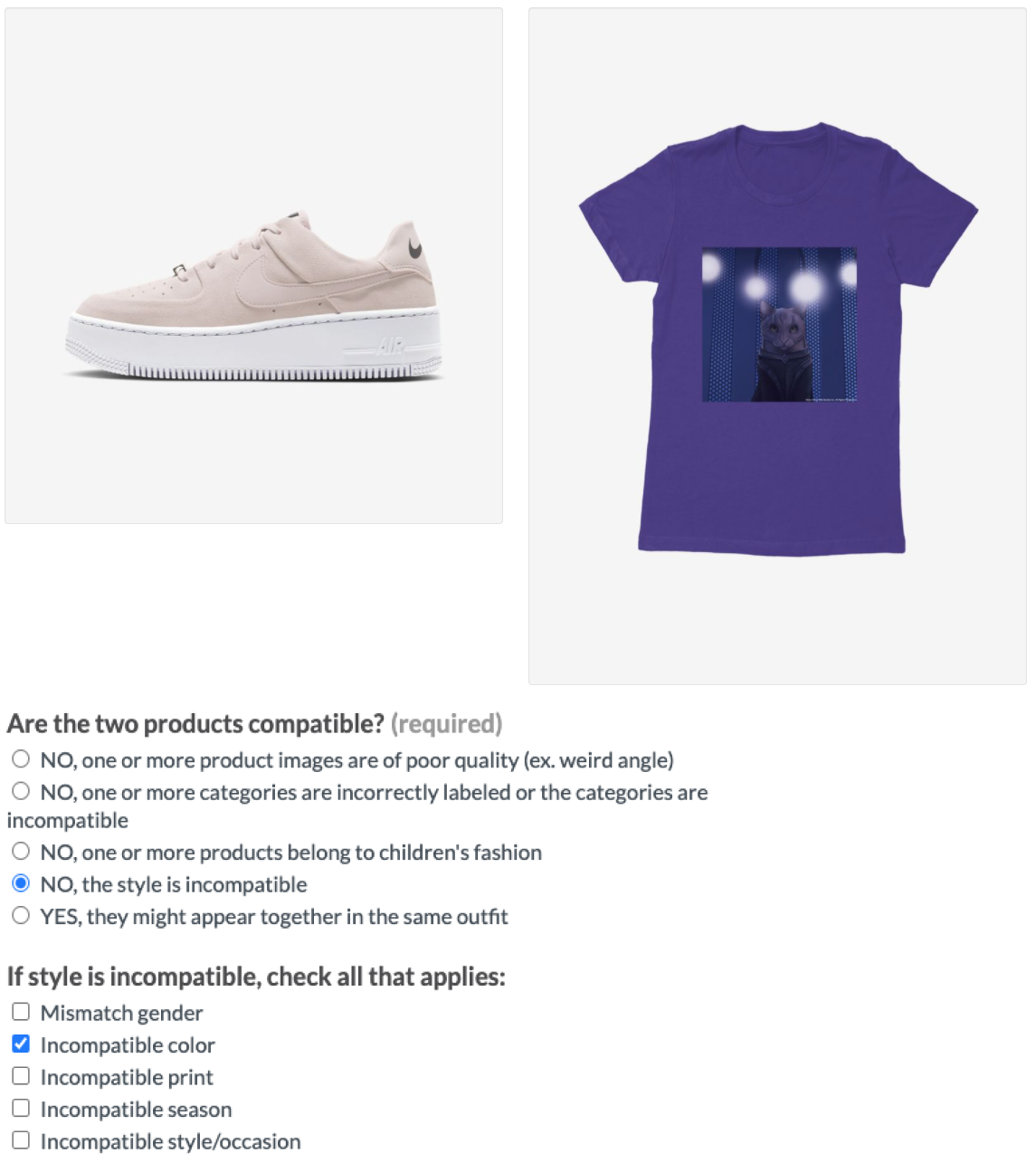}
\caption{Human evaluation task: for each model, we retrieve CTL results for a sampled set of products, and ask whether each $(query, candidate)$ pair is compatible.}
\label{fig:human_eval}
\vspace{-2em}
\end{figure}

Human judgment is extremely valuable because it directly evaluates how our models perform compared with a human stylist. The labeling template we developed is shown in Figure \ref{fig:human_eval}. Since this is a highly specialized task, we decided to only use in-house fashion specialists who follow a comprehensive set of labeling guidelines. These labeling guidelines describe each failure mode (e.g. mismatch color, print, season) with examples and attempt to reduce any subjectivity. We have picked the best performing variant for each of \{Siamese, Triplets, and Classification\} (refer to Section \ref{sec:style_learning}) for comparison.

The questions for this task were generated by sampling 120 products from Pinterest and running the CTL system end-to-end (see Figure \ref{fig:system_overview}). We compute precision by evaluating whether each $(query, candidate)$ pair is a compatible match. In addition, we also asked for qualitative feedback such as common failures and most jarring mistakes. Both quantitative and qualitative feedback can be found in Section \ref{human-eval}.

\subsection{Comparing style extractor training methods}
\label{sec:compare-methods}
We explored several approaches to training the CTL style extractor for the fashion visual complements task (refer to \ref{sec:style_learning}). We report metrics on R@\{1,5,10\} and FITB accuracy using our test set.

\textbf{Scene-based CTL.}
This is the same dataset and model as \cite{ctl-scene2product}, and we include this evaluation as a baseline (See Table~\ref{tab:methods}). This model was trained for the task of retrieving complementary products given a scene image (e.g., the beach). The training uses \textit{(anchor\_scene, positive\_product, negative\_product)} with triplet loss, and a separate network for scene and product images. For this evaluation, we only use the product network since there are no scenes in our dataset. We see that style embeddings trained in this manner do not generalize well to the new task of product-to-products recommendation.

\textbf{Classification with ``proxy'' loss.}
We found that training for the classification task as described in Section \ref{sec:style_learning} showed promising qualitative results, but performed worse than metric learning methods (See Table~\ref{tab:methods}).

\textbf{Siamese.}
We train with contrastive loss. We try the sampling strategy described in \cite{siamese} of 16:1 negative-to-positive ratio, but we find that a simple 1:1 strategy performs better (See Table~\ref{tab:methods}).

\textbf{Triplets.}
We generate triplets $(anchor, positive, negative)$ from our outfits by taking any two items from the same outfit as anchor and positive. We compare performance when sampling negatives randomly or from the same category as positives. We found that triplets training while restricting by category out-performs all other methods (See Table~\ref{tab:methods}).

\begin{table}[htb!]
\begin{center}
\begin{tabular}{||l|r|r|r|r||}
\hline
Method & R@1 & R@5 & R@10 & FITB \\ [0.5ex] 
\hline
CTL(scene-based) & 3.9 & 16.5 & 29.6 & 40.1 \\
Classification & 15.8 & 41.6 & 58.8 & 71.9 \\
Siamese(16:1) & 15.2 & 40.7 & 57.6 & 71.9 \\
Siamese(1:1) & 16.0 & 46.0 & 64.3 & 74.3 \\
Triplets(random) & 18.5 & 49.4 & 67.5 & 77.6 \\
\textbf{Triplets(cat)} & \textbf{20.3} & \textbf{51.4} & \textbf{68.8} & \textbf{78.5} \\
\hline
\end{tabular}
\end{center}
\caption{Recall@{1,5,10} and FITB metrics for different CTL training methods.}
\label{tab:methods}
\vspace{-2.5em}
\end{table}


\subsection{Comparing visual featurizers}
In this section we compare the effects of using Unified Embedding, which has the advantage of additional training on Pinterest data, versus ImageNet for weight initialization of the visual featurizer. These experiments were conducted using triplet loss with category-restricted sampling, which is our best performing model (see Table~\ref{tab:methods}).

We find that pretraining on Pinterest data yields substantial gains relative to using default ImageNet weights, with an R@1 gain from 17.6\% to 20.3\% (see  Table~\ref{tab:exp_initialization}). This suggests that there is a substantial domain shift from ImageNet images to Pinterest images.

\begin{table}[h!]
\begin{center}
\begin{tabular}{||l|r|r|r|r||}
\hline
Method & R@1 & R@5 & R@10 & FITB \\ [0.5ex] 
\hline
ImageNet & 17.6 & 47.2 & 64.0 & 75.5 \\
\textbf{Unified} & \textbf{20.3} & \textbf{51.4} & \textbf{68.8} & \textbf{78.5} \\
\hline
\end{tabular}
\end{center}
\caption{A comparison of ImageNet vs. Unified Embedding weight initialization for the visual featurizer.}
\label{tab:exp_initialization}
\vspace{-2em}
\end{table}

\subsection{Comparing training set sizes}

In this section we compare how decreasing training set size affects performance (see Table~\ref{tab:exp_dataset_size}). This helps us answer the question, ``How much training data is enough?''. We use triplet loss with category-restricted sampling, which is our best performing model in Table~\ref{tab:methods}. Performance on the test set continues to increase as we add training examples, although 10K to 100K outfits sees a significant improvement (R@1 12.1\% -> 16.4\%) while 100K to 1M outfits sees a lesser gain (R@1 16.4\% -> 20.3\%).

\begin{table}[ht!]
\begin{center}
\begin{tabular}{||l|l|r|r|r|r||}
\hline
\#Outfits & \#Objects & R@1 & R@5 & R@10 & FITB \\ [0.5ex] 
\hline
10K & 46K & 12.1 & 37.0 & 54.1 & 66.0 \\
\textbf{100K} & \textbf{453K} & \textbf{16.4} & \textbf{44.3} & \textbf{62.6} & \textbf{74.8} \\
200K & 835K & 16.8 & 46.3 & 64.5 & 75.1 \\
500K & 2.1M & 19.1 & 49.2 & 66.6 & 77.1 \\
\textbf{1M (full)} & \textbf{4.13M (full)} & \textbf{20.3} & \textbf{51.4} & \textbf{68.8} & \textbf{78.5} \\
\hline
\end{tabular}
\end{center}
\caption{An ablation of model performance versus training dataset size. The 100K dataset is released.}
\label{tab:exp_dataset_size}
\vspace{-1.5em}
\end{table}

\subsection{Human evaluation and qualitative feedback}~\label{human-eval}

\begin{figure}[htb!]
\centering
\includegraphics[width=0.8\linewidth]{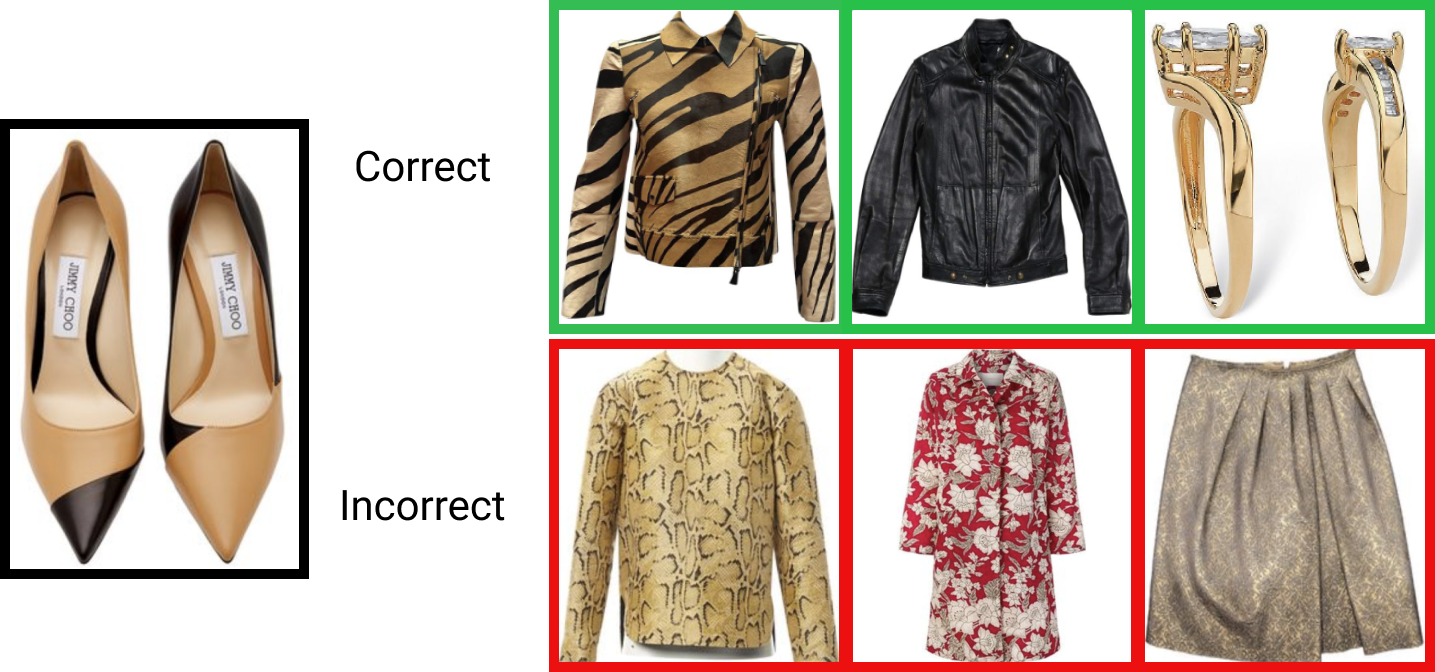}
\caption{Example of correct vs. incorrect evaluation results for a given query, by the standards of a human stylist.}
\label{fig:human_eval_results}
\vspace{-0.5em}
\end{figure}

We asked our in-house fashion specialists to evaluate the end-to-end CTL quality through the eyes of a stylist (refer to Section \ref{sec:eval_human_judgement}). The results are shown in Table~\ref{tab:human_eval_results} and Figure \ref{fig:human_eval_results}. The quantitative results are consistent with our other evaluations, and shows that the triplet loss with category-restricted sampling is the best performing model.

\begin{table}[h!]
\begin{center}
\begin{tabular}{||l|r||}
\hline
Method & Overall Precision \\ [0.5ex] 
\hline
Siamese (1:1) & 47.7  \\
Classification & 50.6   \\
\textbf{Triplets (cat)} & \textbf{55.0}  \\
\hline
\end{tabular}
\end{center}
\caption{Human evaluation results comparing best performing variants from Table \ref{tab:methods}.}
\label{tab:human_eval_results}
\vspace{-2em}
\end{table}

We also asked our in-house specialists for qualitative feedback, and it was clear from their feedback that---when compared with a human stylist---CTL relies more on color and pattern matching. This can result in recommendations that either lack diversity or the ``human touch'' if matches are always the same color or pattern, or are extremely off if the color or pattern does not match entirely. This suggests additional investigation into ensuring that there is sufficient clothing diversity in the training set, as Section \ref{sec:dataset_cleanup} describes.

\begin{figure*}[ht!]
\includegraphics[width=0.9\linewidth]{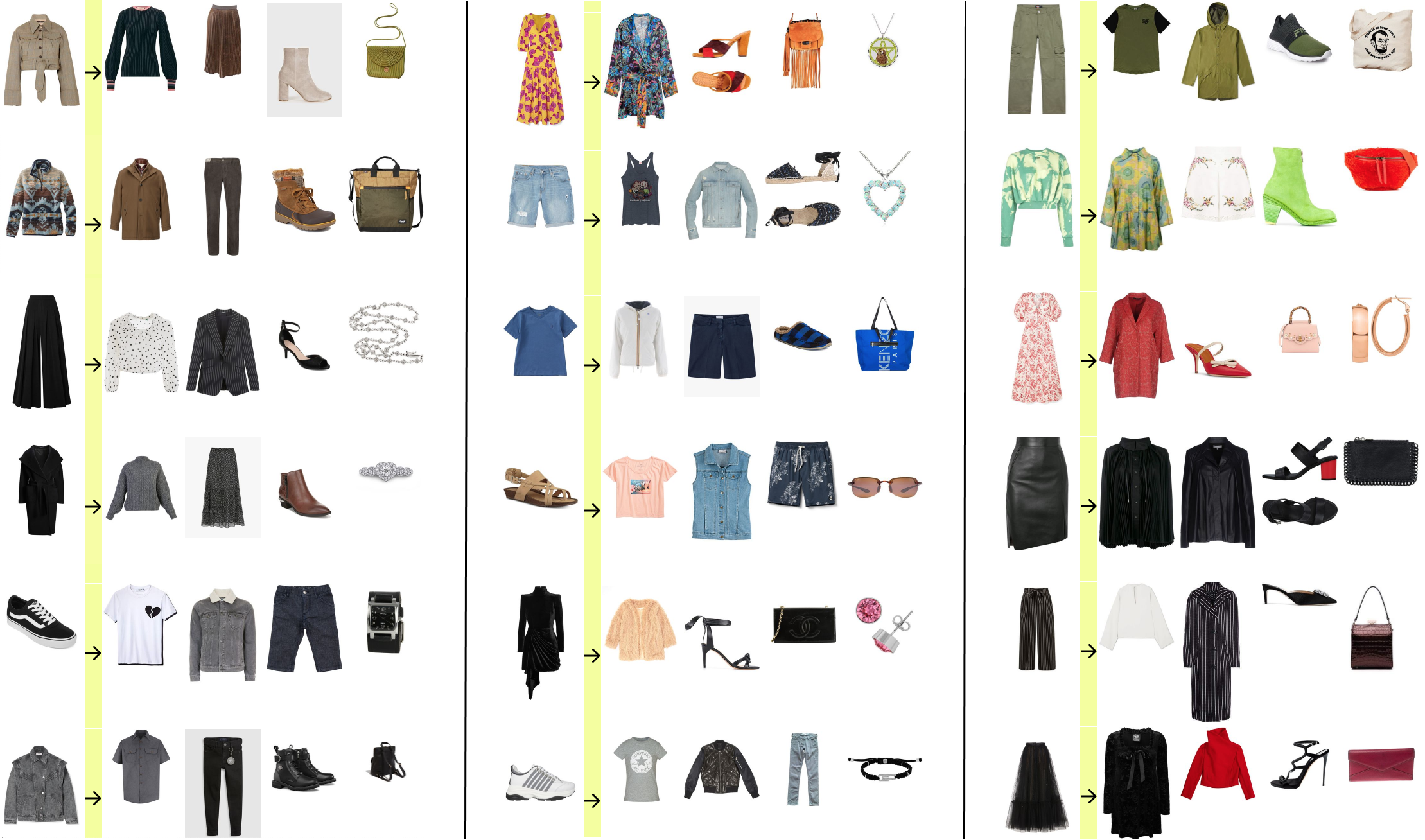}
\caption{Complete The Look retrieval results from Pinterest product corpus. The first column is the query image and the four images to the right are recommendations from different categories.}
\label{fig:qualitative}
\end{figure*}

\subsection{User studies}
As of early 2020, we have a prototype of CTL that is released internally. Qualitative results can be found in Figure \ref{fig:qualitative}. Although we expect more improvements before general launch, this prototype enabled us to conduct user studies to get early feedback. 

We invited 5 Pinterest users to come in for hour-long sessions during which we guided them through using the CTL product feature. Users had high expectations for CTL results, and they wanted to see matches that resembled curation from a human stylist. Overall, the user studies demonstrated that users do see the value of the CTL product, which confirms our continued investment in building it out.
\section{Conclusion and future work}
We shared our approach and results from bootstrapping to productionizing the Complete The Look (CTL) system.

We implemented an automated data pipeline that generates a labeled image dataset for the fashion outfit recommendation task. For our model, we ran comprehensive sets of experiments comparing multiple methods, relying on offline metrics such as R@K that best mirror our product use case. Serving the model in production proved challenging as the problem scaled to tens of millions of unique products. To keep recommendation quality high, we trained and utilized image classifiers that help us trigger CTL only on images that match the training set's domain.

In the future, we hope to add more training data from different image styles (e.g., cropped outfit, full outfit, stock photo) to make the model more robust to domain variations. We also hope to incorporate price as an input, since it is a poor user experience to match a \$8,000 ring with a \$20 dress. Furthermore, we hope to add user personalization, so that we can tailor recommendations to a particular user's style.

\section{Acknowledgments}
The authors would like to thank the rest of the visual search team, especially Chuck Rosenberg for his leadership, Kofi Boakye for his thoughtful editing, and Angela Guo for her feedback and guidance. We would also like to thank Mariellen Barros and Marta Scotto for labeling efforts; Joyce Zha, Claire Li, and Suzy Kim for frontend support; Cherrylyn Cawit and Shilpa Banerjee for conducting user studies.

\bibliographystyle{ACM-Reference-Format}
\bibliography{references}

\end{document}